\documentclass[12pt]{article}
\usepackage{algorithm}
\usepackage{algorithmic}
\usepackage{latexsym}
\usepackage{cite}
\usepackage{enumitem}
\usepackage{amssymb,mathrsfs}
\usepackage{booktabs}
\usepackage{subfigure}
\usepackage{float}
\usepackage{color}
\usepackage{amsmath}
\usepackage[colorlinks,
            linkcolor=blue,
            anchorcolor=red,
            citecolor=blue]{hyperref}
\usepackage{bm}
\usepackage{natbib}
\usepackage{graphicx}
\usepackage{float}
\usepackage{subfigure}
\usepackage{wrapfig}
\usepackage{amsthm}
\usepackage{epstopdf}
\usepackage{lscape}
\usepackage{multirow}
\usepackage[table,xcdraw]{xcolor}
\usepackage{booktabs}
\usepackage{pifont}

\setcitestyle{round}
\allowdisplaybreaks[4]

\def \a{\rm Adam}
\def \e{\rm EAdam}

\begin{document}
\title{EAdam Optimizer: How $\epsilon$ Impact Adam}

\author {
        Wei Yuan, 
        Kai-Xin Gao\\
        {\small School of Mathematics, Tianjin University \vspace{-0.5em}}\\
        {\small Email: yuann@tju.edu.cn, gaokaixin@tju.edu.cn \vspace{-0.5em}}
}

\date{}

\maketitle

\begin{abstract}
\noindent
Many adaptive optimization methods have been proposed and used in deep learning, in which Adam is regarded as the default algorithm and widely used in many deep learning frameworks. Recently, many variants of Adam, such as Adabound, RAdam and Adabelief,  have been proposed and show better performance than Adam. However, these variants mainly focus on changing the stepsize by making differences on the gradient or the square of it. Motivated by the fact that suitable damping is important for the success of powerful second-order optimizers, we discuss the impact of the constant $\epsilon$ for Adam in this paper. Surprisingly, we can obtain better performance than Adam simply changing the position of $\epsilon$. Based on this finding, we propose a new variant of Adam called EAdam, which doesn't need extra hyper-parameters or computational costs. We also discuss the relationships and differences between our method and Adam. Finally, we conduct extensive experiments on various popular tasks and models. Experimental results show that our method can bring significant improvement compared with Adam. Our code is available at https://github.com/yuanwei2019/EAdam-optimizer.
\vspace{3mm}

\vspace{3mm}


\end{abstract}

\section{Introduction}
Deep neural network models have shown state-of-the-art performance in many application areas, such as image classification \citep{net2016}, natural language processing \citep{nlp2015} and object detection \citep{oc2016}. As models becoming more and more complex, finding efficient training methods has attracted many researchers.

Among the algorithms for training deep networks, stochastic gradient descent (SGD) \citep{sgd} is one of the most widely used algorithms, due to its ease of implementation and low computational costs. However, SGD scales the gradient uniformly in all directions, which may lead to relatively-slow convergence and sensitivity to the learning rate \citep{adabound2019,adamod2019}. In order to solve this problems, several adaptive methods have been proposed, in which the adaptive moment estimation of the past squared gradients is computed to scale the gradient of different parameters. Examples of these adaptive methods include AdaGrad \citep{ada2011}, AdaDelta \citep{ada2012}, RMSprop \citep{rms2012}, and Adam \citep{ada2014}. In particular, Adam has been regarded as the default algorithm and widely used in many deep learning frameworks \citep{wil2017}.

Although Adam is popular in many areas, its generalization ability is likely worse than the SGD family (non-adaptive methods) \citep{wil2017}. What's more, \citet{conadam2018} focused on the non-convergence in Adam and elucidated how the exponential moving average causes it. So, many variants of Adam have been proposed to solve these problems, such as AMSGrad \citep{conadam2018}, MSVAG \citep{msvag2017}, RAdam \citep{radam2019}, AdaBound \citep{adabound2019} and Adabelief \citep{adabelief2020}. These modifications usually have better performance compared to Adam.

As defined in \citet{ada2014}, let $g_t$ denote the gradient at the $t$-th timestep, $m_t, v_t$ denote the exponential moving averages of $g_t, g_t^2$, respectively and $\alpha$ denote the stepsize. Then the adaptive stepsize in Adam can be seen as $\alpha/(\sqrt{v_t}+\epsilon)$, where $\epsilon$ is a constant with default value $10^{-8}$. More details are given in Algorithm \ref{adam}. Summarizing the variants of Adam, we find that most of these methods are focused on $v_t$. They all scale the factor $m_t$ by making different changes to $v_t$ and usually ignore the impact of $\epsilon$. The reason why adds $\epsilon$ to $\sqrt{v_t}$ is to keep the denominator from getting too small. It's similar to the damping factor in second order optimization methods. However, powerful second-order optimizers (such as Hessian-Free optimization approach \citep{hf2013} and natural gradient descent \citep{kfac2015}) usually require more sophisticated damping solutions to work well, and will may be completely fail without them \citep{kfac2015}.
Therefore, a direct question is whether $\epsilon$ has a great influence on the performance of Adam.

Motivated by the above problem, we explored whether changing the update rule of $\epsilon$ in Adam would obtain better performance. An interesting finding is that we can obtain better performance than Adam simply changing the position of $\epsilon$ without extra hyper-parameters and computational costs. In particular, we make the
following key contributions:

\begin{itemize}
  \setlength{\itemsep}{3pt}
  \setlength{\parsep}{0pt}
  \setlength{\parskip}{0pt}
  \item Based on the above finding, we propose a new variant of Adam, called $\epsilon$-Adam (EAdam), in which $\epsilon$ is added to $v_t$ at every step before the bias correction and it is accumulated through the updating process (details are given in Section \ref{sec-3}). We also discuss the difference of EAdam compared with Adam and show that Adam can not achieve same performance as EAdam by tuning $\epsilon$ directly in experiments.
  \item We turn to an empirical study of EAdam on various popular tasks and models in image classification, natural language processing and object detection compared with Adam, RAdam and Adabelief. Experimental results demonstrate that our method outperforms Adam, and RAdam and has similar performance to Adabelief in all tasks and models. Compared these variants of Adam, EAdam can achieve better or similar results with simpler variation to Adam.
\end{itemize}

\begin{algorithm}[htb]
\caption{Adam algorithm}\label{adam}

\begin{algorithmic}
\REQUIRE $f(\theta)$: the objective function with parameters $\theta$
\REQUIRE $\alpha$: stepsize
\REQUIRE $\beta_1,\beta_2$: exponential decay rates
\REQUIRE $\theta_0$: initial parameters

\STATE  $m_0\leftarrow0, v_0\leftarrow0, t=0$

\WHILE{convergence is not reached}

\STATE $t\leftarrow t+1$
\STATE $g_t\leftarrow \nabla_\theta f_t(\theta_{t-1})$
\STATE $m_t\leftarrow \beta_1m_{t-1}+(1-\beta_1)g_t$
\STATE $v_t\leftarrow \beta_2v_{t-1}+(1-\beta_2)g_t^2$
\STATE $\hat{m}_t= m_t/(1-\beta_1^t)$, $\hat{v}_t=v_t/(1-\beta_2^t)$
\STATE $\theta_t\leftarrow \theta_{t-1}-\alpha(\hat{m}_t/\sqrt{\hat{v}_t}+\epsilon)$

\ENDWHILE

\end{algorithmic}
\end{algorithm}

\section{Related work}

There have been many optimizers proposed to train deep networks. Firstly, we summarize a few closely related works of the variants of Adam. MSVAG \citep{msvag2017} combines taking signs and variance adaptation into Adam. AMSGrad \citep{conadam2018} resolves the non-convergence in Adam based on the long-term memory of past gradients. RAdam \citep{radam2019} analyses the convergence issue in Adam and explicitly rectifies the variance of the learning rate. AdaBound \citep{adabound2019} employs dynamic bounds on learning rate in adaptive methods and clips the extreme learning rate. Adabelief \citep{adabelief2020} considers to adapt the stepsize according to the belief in the current gradient direction and obtains good generalization performance. Of course, there are many other variants of Adam (such as NAdam \citep{nadam2016}, AdamW \citep{adamw2017}, AdaMod \citep{adamod2019} and TAdam \citep{tadam2020}), and we will not introduce birefly here.

Besides first-order methods, many optimization algorithms, such as quasi-Newton method (\citeauthor{qn2011}, \citeyear{qn2011}; \citeauthor{qn2019}, \citeyear{qn2019}), Hessian-Free optimization approach (\citeauthor{hf2010}, \citeyear{hf2010}; \citeauthor{hf2017}, \citeyear{hf2017}) and Kronecker-factored approximate curvature (KFAC) (\citeauthor{kfac2015}, \citeyear{kfac2015}; \citeauthor{kfc2016}, \citeyear{kfc2016}), have been presented and used for training deep networks. Second-order optimization algorithms can greatly accelerate convergence by using curvature matrix to correct gradient. However, the computation of curvature matrix and its inverse leads to heavy computational burden, and hence second-order optimizers are not widely used in deep learning.

\section{Method}\label{sec-3}
\subsection{Algorithm}
{\bf Notations: }  In this paper, we will use the following basic notations as defined in \citep{ada2014}.
\begin{itemize}
  \setlength{\itemsep}{3pt}
  \setlength{\parsep}{0pt}
  \setlength{\parskip}{0pt}
  \item $f(\theta)$: the objective function with parameters $\theta$
  \item $g_t=\nabla_\theta f(\theta)$: the gradient of $f(\theta)$ at the $t$-th step
  \item $\alpha$: stepsize (learning rate), the default setting is 0.001
  \item $m_t, v_t$: exponential moving average of $g_t, g_t^2$, respectively
  \item $\beta_1,\beta_2 \in(0,1)$: the exponential decay rates of moving
averages, typically set as $\beta_1=0.9$ and $\beta_2=0.999$
  \item $\epsilon$: a small positive constant, and the typical value is $10^{-8}$
\end{itemize}

\noindent {\bf EAdam: } The Adam algorithm and our algorithm are summarized in Algorithm \ref{adam} and Algorithm \ref{eadam} respectively, in which the differences are marked in blue and the operations for vectors are element-wise. Our method mainly change the update rule of $\epsilon$, so we call this proposed method \emph{EAdam}. For the standard Adam, $\epsilon$ is added to the factor $\sqrt{\hat{v}_t}$ after the bias correction step, that is $\sqrt{\hat{v}_t}+\epsilon$. So the adaptive stepsize in Adam is $\alpha/(\sqrt{\hat{v}_t}+\epsilon)$. For EAdam, $\epsilon$ is added to $v_t$ at every step before the bias correction, it is accumulated through the updating process, and finally the adaptive stepsize in EAdam is $\alpha/\sqrt{\hat{v}_t}$. Compared with Adam, the change in our method is simple but it is efficient without extra computation cost and hyper-parameters setting. More discussion is given in the following subsection. What's more, EAdam shows better performance compared with Adam in experiments (results are given in Section \ref{sec-5}).

\begin{algorithm}[htb]
\caption{EAdam algorithm. This algorithm is proposed based on the Adam algorithm and the differences from standard Adam are shown in {\color{blue}blue}. The default settings are same as Adam, where $\beta_1=0.9, \beta_2=0.999, \epsilon=10^{-8}.$}\label{eadam}

\begin{algorithmic}
\REQUIRE $f(\theta)$: the objective function with parameters $\theta$
\REQUIRE $\alpha$: stepsize
\REQUIRE $\beta_1,\beta_2$: exponential decay rates
\REQUIRE $\theta_0$: initial parameters

\STATE  $m_0\leftarrow0, v_0\leftarrow0, t=0$

\WHILE{convergence is not reached}

\STATE $t\leftarrow t+1$
\STATE $g_t\leftarrow \nabla_\theta f_t(\theta_{t-1})$
\STATE $m_t\leftarrow \beta_1m_{t-1}+(1-\beta_1)g_t$
\STATE $v_t\leftarrow \beta_2v_{t-1}+(1-\beta_2)g_t^2$
\STATE {\color{blue}$v_t \leftarrow v_t+\epsilon$}
\STATE $\hat{m}_t= m_t/(1-\beta_1^t)$, $\hat{v}_t=v_t/(1-\beta_2^t)$
\STATE $\theta_t\leftarrow \theta_{t-1}-\alpha(\hat{m}_t/{\color{blue}\sqrt{\hat{v}_t}})$

\ENDWHILE

\end{algorithmic}
\end{algorithm}

\subsection{Comparison with Adam}
We will here analyse how the change in EAdam affects the update of parameters $\theta$. Let $g_t$ be the gradient of the objective function at timestep $t$. According to update formulas in Algorithm \ref{adam} and Algorithm \ref{eadam}, $v_t^{(\a)}$ and $v_t^{(\e)}$ can be expressed by the gradients at all previous timesteps as follows
\begin{equation*}
  v_t^{(\a)}=(1-\beta_2)\sum_{i=1}^{t}\beta_2^{t-i}g_i^2,
\end{equation*}
\begin{equation*}
  v_t^{(\e)}=(1-\beta_2)\sum_{i=1}^{t}\beta_2^{t-i}g_i^2+\frac{1-\beta_2^t}{1-\beta_2}\epsilon.
\end{equation*}
After the bias correction step, we have
\begin{equation*}
  \hat{v}_t^{(\a)}=\frac{1-\beta_2}{1-\beta_2^t}\sum_{i=1}^{t}\beta_2^{t-i}g_i^2=G_t,
\end{equation*}
\begin{equation*}
  \hat{v}_t^{(\e)}=\frac{1-\beta_2}{1-\beta_2^t}\sum_{i=1}^{t}\beta_2^{t-i}g_i^2
  +\frac{1}{1-\beta_2}\epsilon=G_t+\frac{1}{1-\beta_2}\epsilon,
\end{equation*}
where $G_t=\frac{1-\beta_2}{1-\beta_2^t}\sum_{i=1}^{t}\beta_2^{t-i}g_i^2$, $\epsilon$ is a small positive constant in EAdam. Then the adaptive stepsize in Adam and EAdam can be seen as
\begin{equation}\label{direction}
\frac{\alpha}{\sqrt{G_t}+\epsilon^{'}}, \quad \frac{\alpha}{\sqrt{G_t+\epsilon/(1-\beta_2)}}
\end{equation}
where $\epsilon^{'}$ is the small positive constant in Adam.

\begin{figure}[htb]
  \centering
  \includegraphics[width=0.4\linewidth]{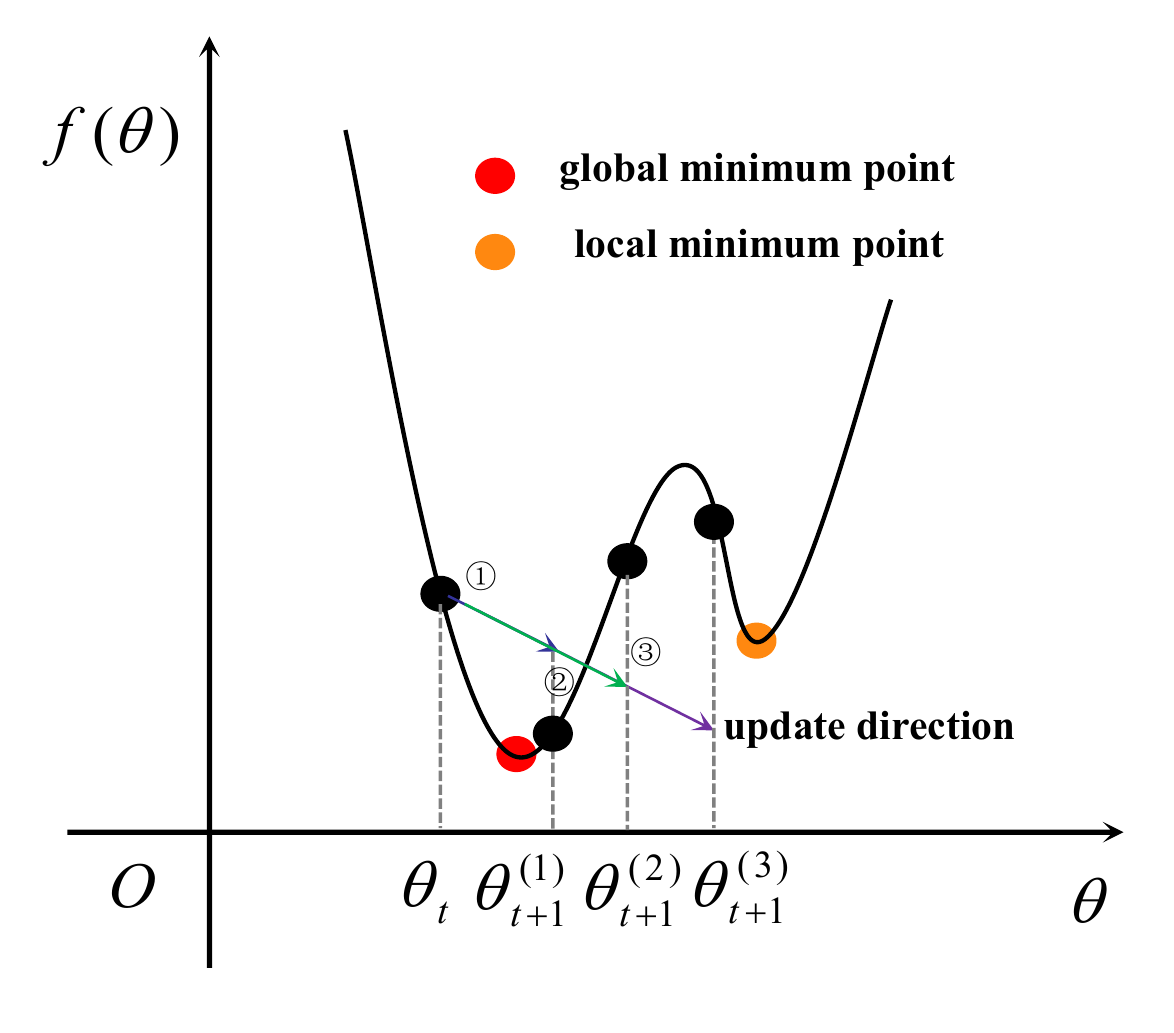}
  \caption{ An intuitive comparison for different stepsizes when $\theta_t$ is close to the global minimum point. We enlarge the gap for visual effect. In fact, $\theta_t$ may be very close to the global minimum point. }\label{com}
\end{figure}

We firstly let $\epsilon^{'}=\epsilon=10^{-8}$, then we want to analyse the differences of stepsizes when using Adam and EAdam to train deep networks. At the begin of training, the elements in $G_t$ are far larger than $\epsilon^{'}$ and $\epsilon$, the stepsizes in Adam and EAdam can all approximated as $\alpha/\sqrt{G_t}$. In this case, the stepsize is determined by $G_t$. Then, the elements in $G_t$ may become small and $\epsilon^{'}$ or $\epsilon$ can affect the elements in $G_t$. In this case, the stepsize is determined by $G_t$ and $\epsilon^{'}$ ($\epsilon$). It easy to see that this case happens earlier in EAdam because $\epsilon$ is added to $G_t$ rather than $\sqrt{G_t}$. Finally, the elements in $G_t$ may become far smaller than $\epsilon^{'}$ or $\epsilon$, and the stepsizes become
\[
\frac{\alpha}{\epsilon^{'}}, \quad
\frac{\alpha}{\sqrt{\epsilon/(1-\beta_2)}}
\]
in Adam and EAdam, respectively. In this case, EAdam takes smaller stepsize than Adam. A possible intuitive comparison is given in Figure \ref{com}. When $\theta_t$ is close to the global minimum point, EAdam takes smaller step (as shown by \ding{172}) can avoid the cases as shown by \ding{173} and \ding{174}. This may be an explanation why EAdam achieves better performance than Adam.

From Eq. (\ref{direction}), we can see that EAdam essentially adds a constant times of $\epsilon$ to $G_t$ before the square root operation. However, this operation is not equivalent to adding a fixed constant $\epsilon^{'}$ to $\sqrt{G_t}$. In other words, we can't find a fixed constant $\epsilon^{'}$ such that $\sqrt{G_t}+\epsilon^{'}=\sqrt{G_t+\epsilon/(1-\beta_2)}$, where $\epsilon$ is known, for the following reasons.

If we let $\sqrt{G_t}+\epsilon^{'}=\sqrt{G_t+\epsilon/(1-\beta_2)}$ where $\epsilon$ is known. Then, we have
\begin{equation}\label{solution}
\epsilon^{'}=\sqrt{G_t+\epsilon(1-\beta_2)}-\sqrt{G_t}.
\end{equation}
Because $G_t$ is constantly updated, $\epsilon^{'}$ is also adjusted based on $G_t$ in the iterative process. Therefore, $\epsilon^{'}$ is not fixed. From this interpretation, the change in EAdam can be seen as adopting an adaptive $\epsilon$ rather than a constant in Adam.

To sum up, we give some intuitive comparisons and explanations for EAdam in this subsection. However, analyzing the reasons why EAdam performances better in theory may be difficult and it is worthy to be further studied.

\section{Experiments}\label{sec-5}

In this section, we perform a thorough evaluation of our EAdam optimizer
against popular and latest optimization methods including Adam \citep{ada2014}, RAdam \citep{radam2019} and Adabelief \citep{adabelief2020} on different deep learning tasks. We focus on these following tasks: image classification
on CIFAR-10 and CIFAR-100 \citep{data2009} with VGG11 \citep{net2014}, ResNet18 \citep{net2016} and DenseNet121 \citep{des2016}, language modeling on Penn Treebank \citep{data1993} with LSTM \citep{lstm2015} and object detection on PASCAL VOC \citep{data2010} with Faster-RCNN + FPN \citep{ob2015,ob2017}. The setup for each task is summarized in Table \ref{tabdata}. 
We mainly use the official implementation of AdaBelief and refer to the hyper-parameter tuning in \citet{adabelief2020} for image classification and language modeling. We mainly refer to the setting in \citep{gc2020} for object detection. 
Experimental results are given in the following subsection.

\begin{table}[htb]
\setlength{\abovecaptionskip}{0.0cm}
\setlength{\belowcaptionskip}{0.3cm}
\centering
\caption{Summaries of the models used in our experiments.}
\label{tabdata}
\begin{tabular}{@{}c|c|c@{}}
\toprule
\textbf{Dataset} & \textbf{Network Type} & \textbf{Architecture} \\ \midrule
CIFAR-10         & Deep Convolutional               & VGG16, ResNet18, DenseNet121                         \\
CIFAR-100        & Deep Convolutional              & VGG16, ResNet18, DenseNet121                        \\
Penn Treebank           & Recurrent               & 1,2,3-Layer LSTM                 \\
PASCAL VOC           & Deep Convolutional              & Faster-RCNN + FPN                 \\
\bottomrule
\end{tabular}
\end{table}

\subsection{Image classification}

\begin{figure}[htb]
  \centering
  \subfigure[VGG11 on CIFAR-10]{
  \includegraphics[width=0.31\linewidth]{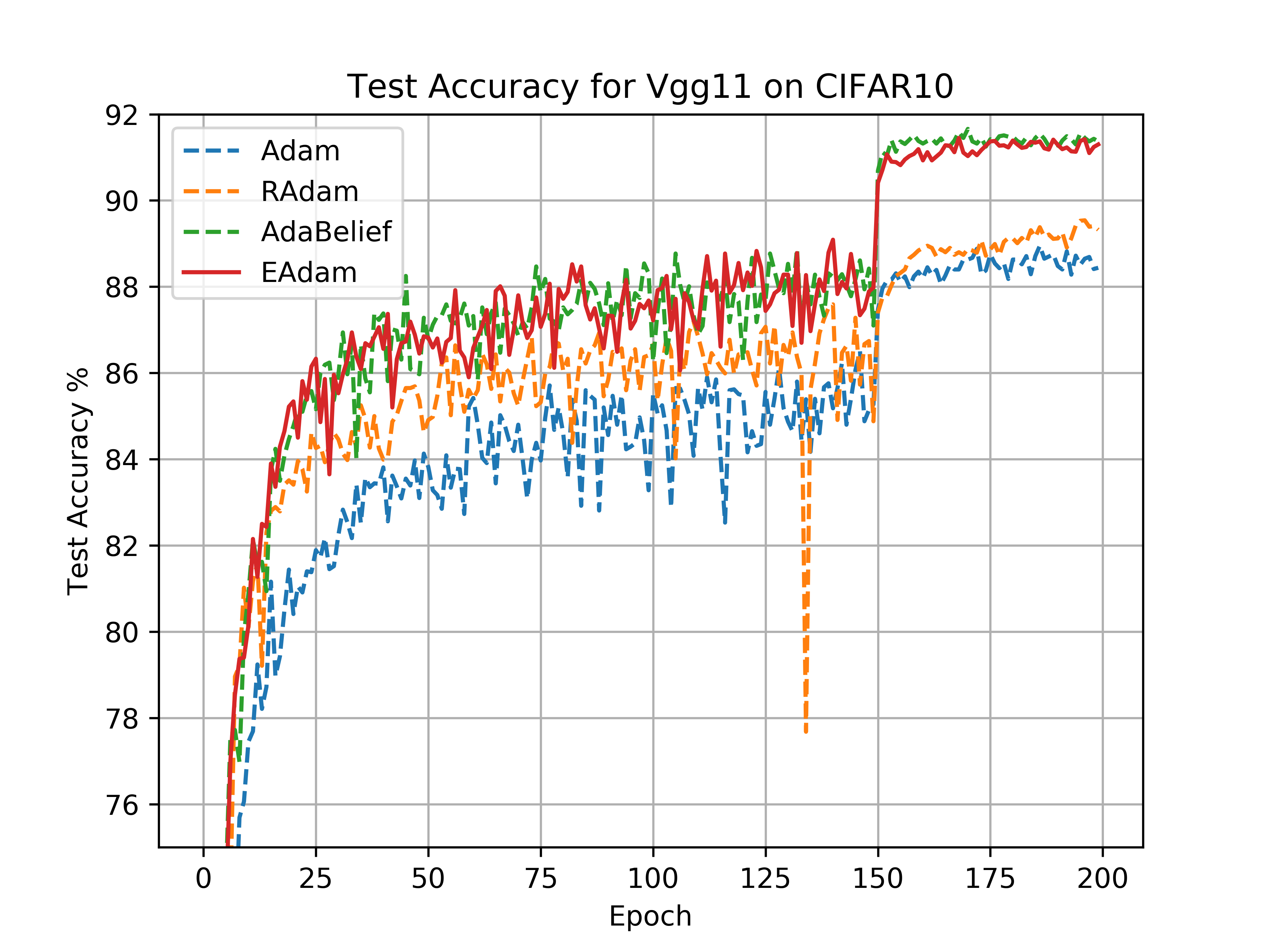}}
  \subfigure[ResNet18 on CIFAR-10]{
  \includegraphics[width=0.31\linewidth]{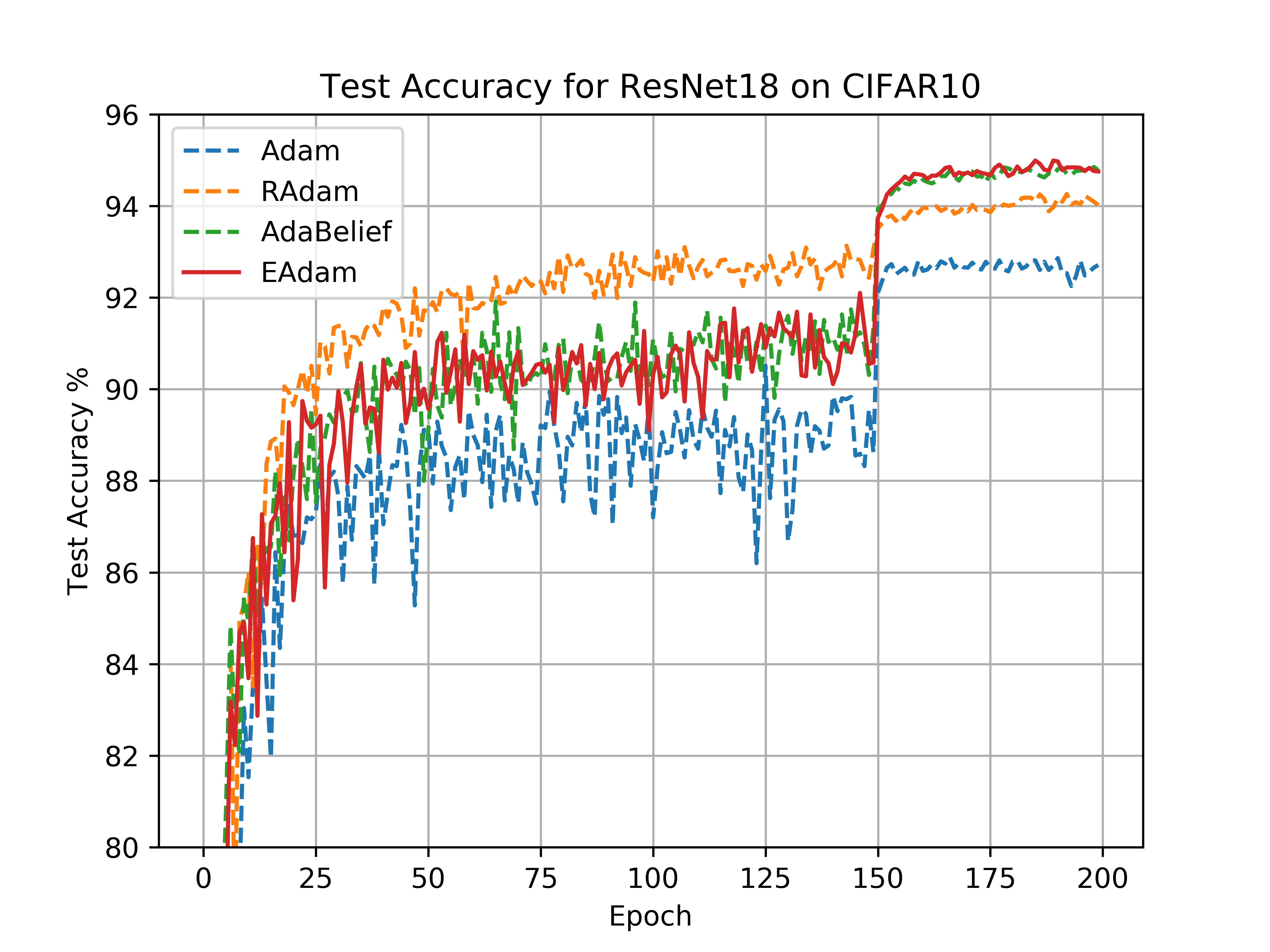}}
  \subfigure[DenseNet121 on CIFAR-10]{
  \includegraphics[width=0.31\linewidth]{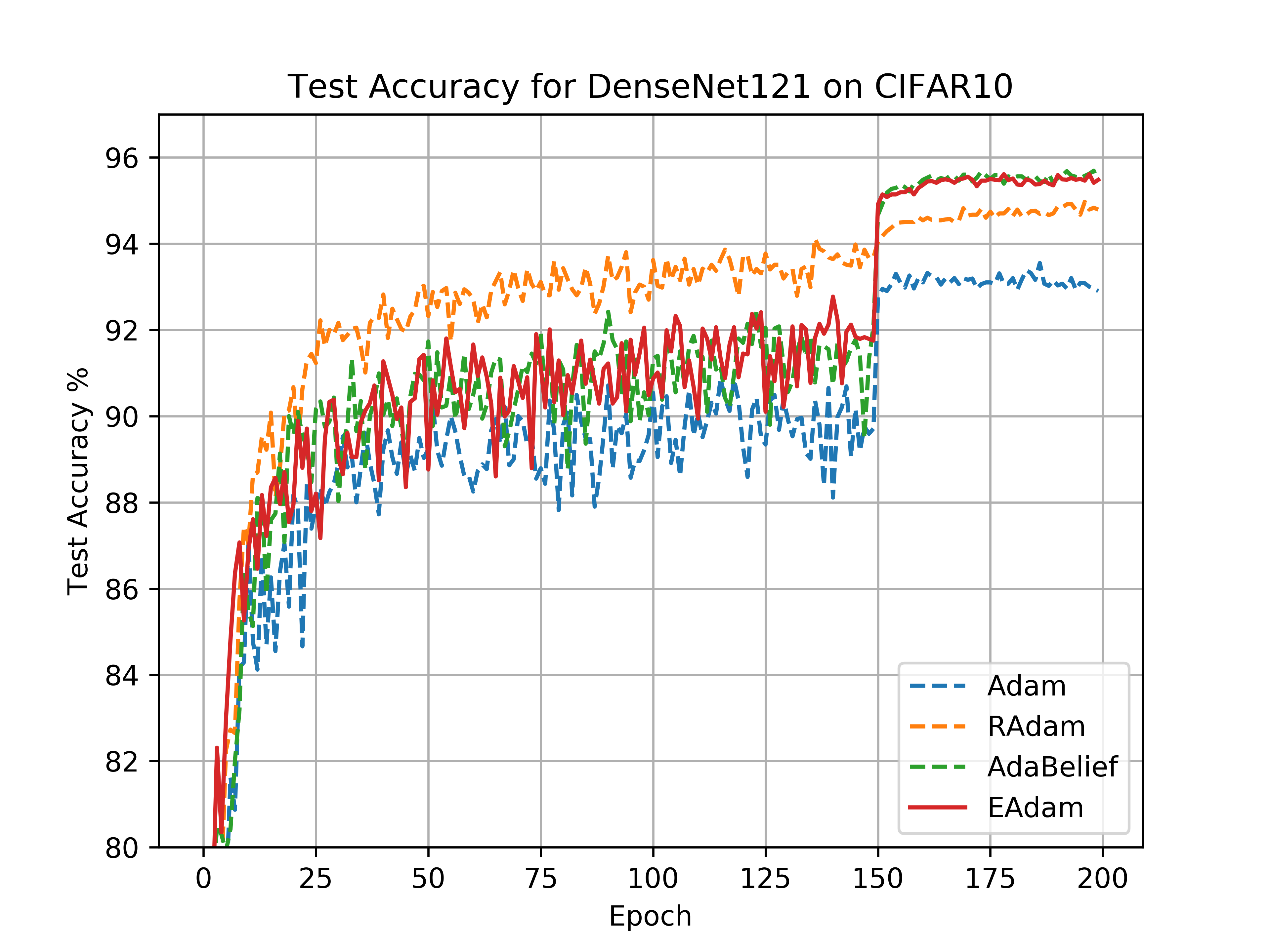}}\\
  \subfigure[VGG11 on CIFAR-100]{
  \includegraphics[width=0.31\linewidth]{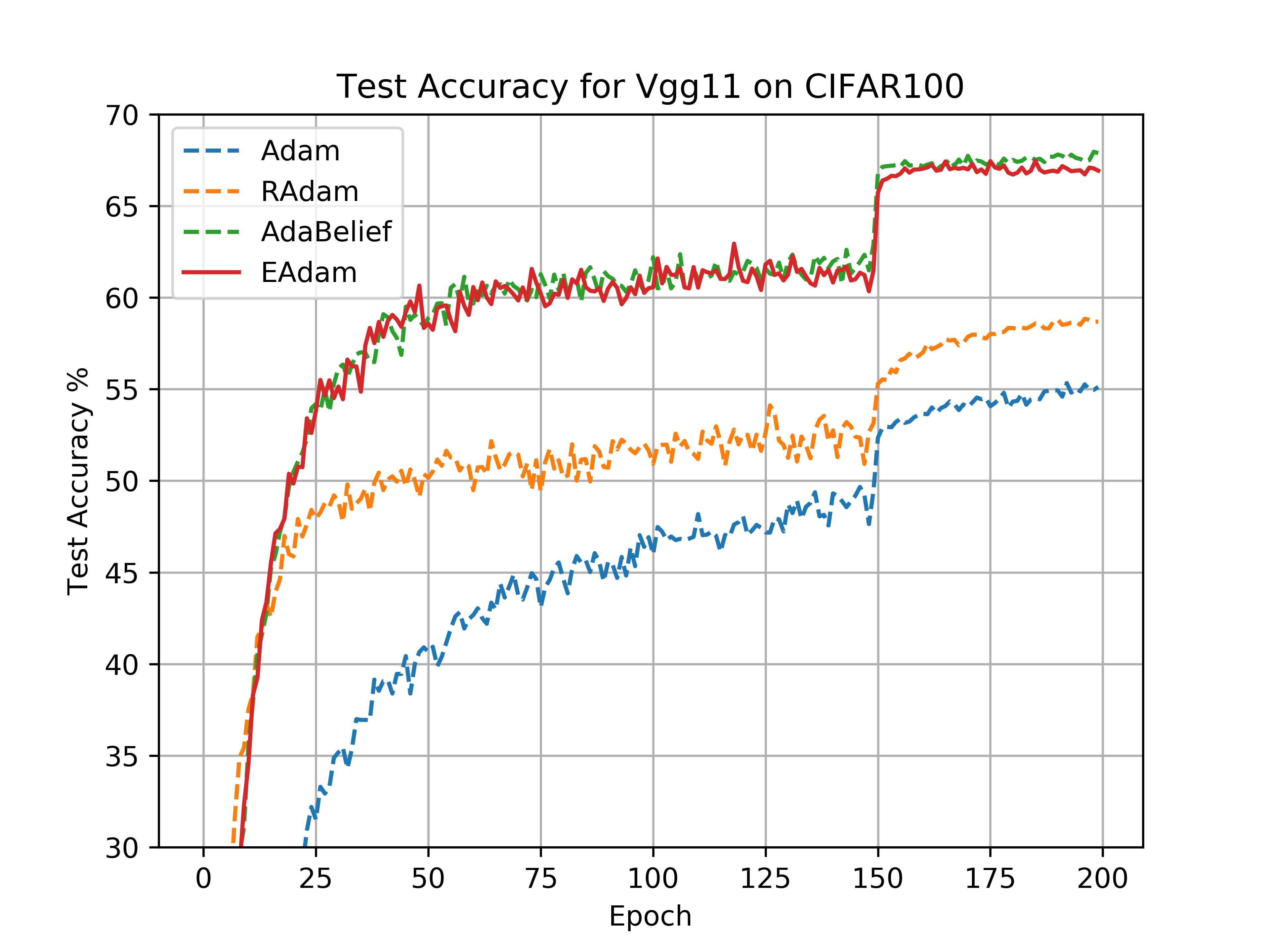}}
  \subfigure[ResNet18 on CIFAR-100]{
  \includegraphics[width=0.31\linewidth]{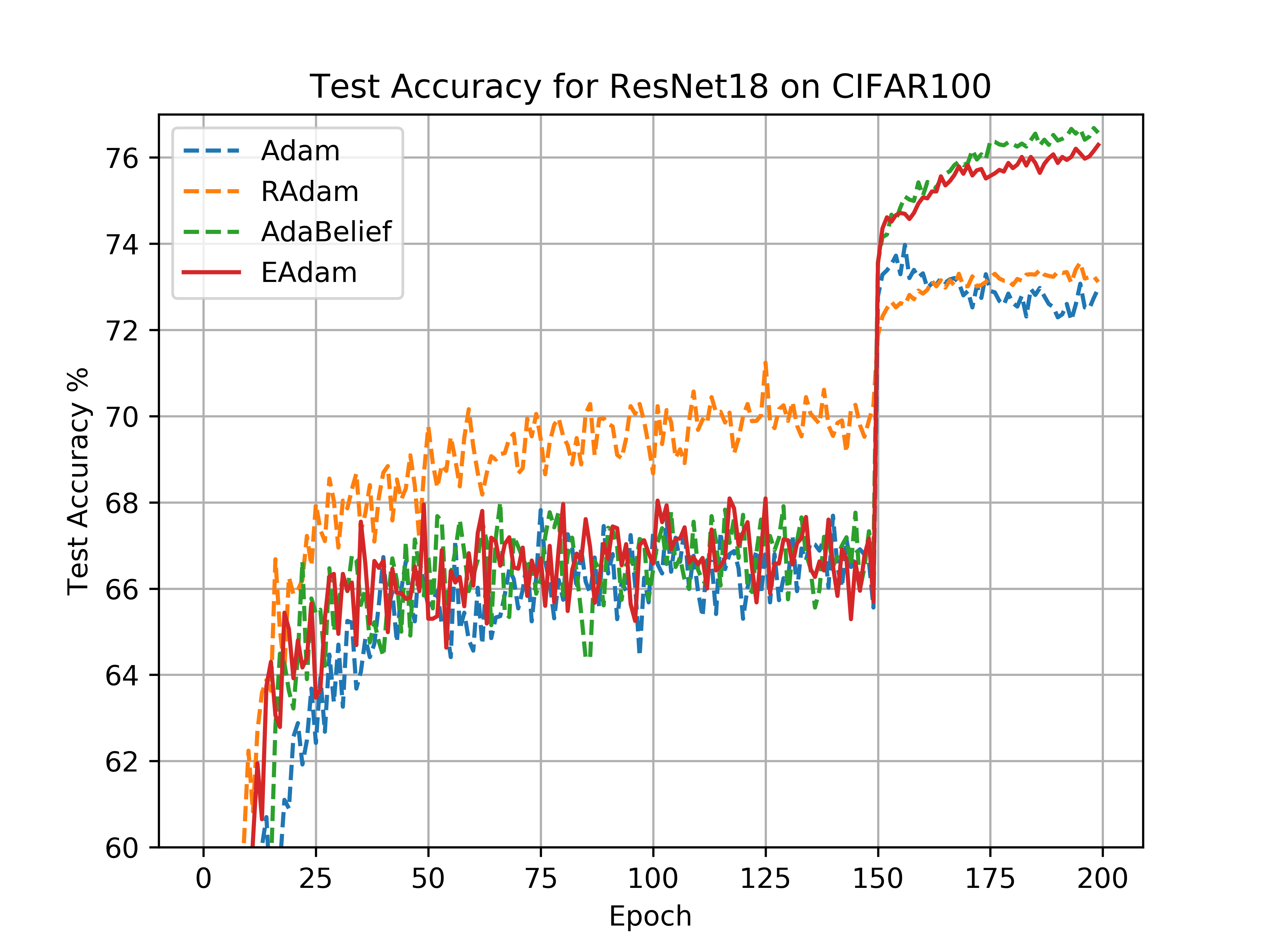}}
  \subfigure[DenseNet121 on CIFAR-100]{
  \includegraphics[width=0.31\linewidth]{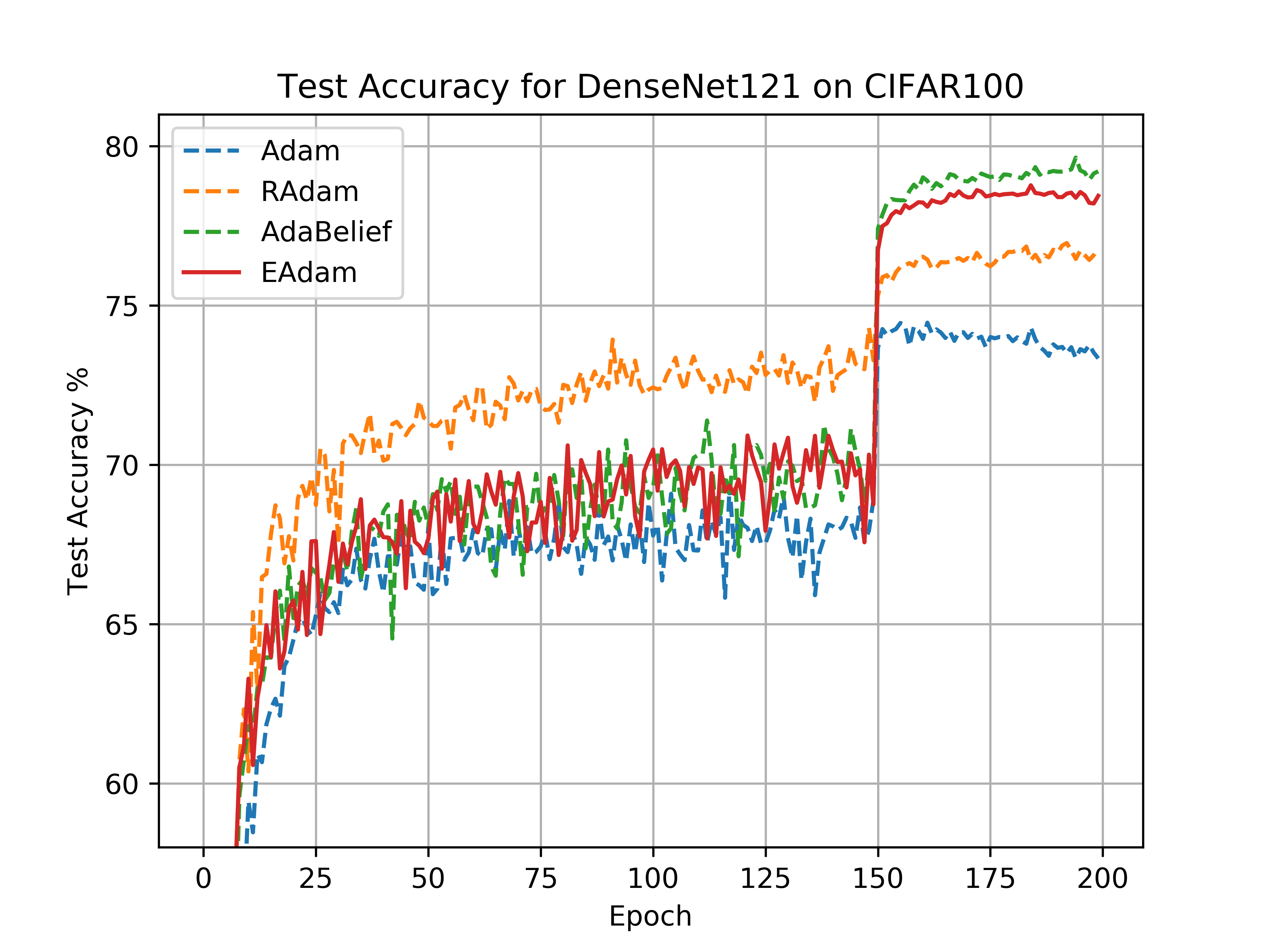}}
  \caption{The curves of testing accuracy with epochs for Adam, RAdam, Adabelief and EAdam on CIFAR-10 and CIFAR-100. The models we used here are VGG11, ResNet18 and DenseNet121.}\label{picacc} 
\end{figure}

Firstly, we consider the image classification task on CIFAR-10 and CIFAR-100 datasets. For all experiments in this part, we all use the default parameters of Adam for all methods ($\alpha=10^{-3},\beta_1=0.9, \beta_2=0.999, \epsilon=10^{-8}$). We set the weight decay as $5e-4$ of all methods. We train all models for 200 epochs with batch-size 128 and multiply the learning rates by 0.1 at the 150-th epoch. Results are summarized in Figure \ref{picacc} and Table \ref{tabacc}.

\begin{table}[htb]
\small
\setlength{\abovecaptionskip}{0.0cm}
\setlength{\belowcaptionskip}{0.3cm}
\centering
\caption{Test accuracy for VGG11 ResNet18 and DenseNet121 on CIFAR-10 and CIFAR-100. }
\label{tabacc}
\begin{tabular}{@{}cc|ccccc@{}}
\toprule
Dataset   & Model    & Adam            & RAdam            & Adabelief           & EAdam           \\ \midrule
CIFAR-10      & VGG11    & 88.95 & 89.54 & 91.66 & 91.45  \\
CIFAR-10      & ResNet18 & 92.88 & 94.26 & 94.85 & 94.99  \\
CIFAR-10  & DenseNet121   & 93.55 & 94.97 & 95.69 & 95.61  \\
CIFAR-100  & VGG11 & 55.33 & 58.84 & 67.88 & 67.45  \\
CIFAR-100 & ResNet18    & 73.79 & 73.56 & 76.68 & 76.29 \\
CIFAR-100 & DenseNet121 & 74.46 & 76.96 & 79.64 & 78.77 \\ \bottomrule
\end{tabular}
\end{table}

{\bf VGG11}:
The accuracy curves on VGG11 are shown in the first column of Figure \ref{picacc}. We can see that EAdam outperforms Adam and RAdam and has similar performance to Adabelief on both datasets. Compared with Adam, our method has both faster convergence and better performance. As shown in Table \ref{tabacc}, the test accuracy of EAdam is improved about $2.5\%$ and $12\%$ than Adam on CIFAR-10 and CIFAR-100, respectively. It is also obvious that EAdam outperforms RAdam. AdaBelief is a latest variant of Adam and shows good generalization as in the SGD family with fast convergence as adaptive methods. EAdam can keep similar performance to it without change to $g_t^2$ in Adam.

{\bf ResNet18}:
The accuracy curves on ResNet18 are shown in the second column of Figure \ref{picacc}. As we except, EAdam also has both faster convergence and higher test accuracy than Adam. EAdam achieves about approximately $2\%$ and $2.5\%$ improvement on CIFAR-10 and CIFAR-100 in terms of test accuracy as shown in Table \ref{tabacc}. Compared with RAdam, EAdam converges slower than RAdam in the early training. But when the learning rates are decayed at the 150-th epoch, EAdam begins to outperform RAdam. On the CIFAR-10 dataset, EAdam has similar performance to AdaBelief. On the CIFAR-100 dataset, EAdam has similar  convergence to AdaBelief but the testing accuracy is slightly lower.

{\bf DenseNet121}:
The accuracy curves on ResNet18 are shown in the third column of Figure \ref{picacc}. We can see that the overall performance of each method on DenseNet121 is similar to that on ResNet18. On the CIFAR-100 dataset, the improvement of EAdam relative to Adam becomes more significant than that on ResNet18, which is enhanced with more than $4\%$ in the test accuracy. In addition, the testing accuracy of Adam reduces about $0.9\%$ than Adabelief on the CIFAR-100 dataset. To sum up, the change of $\epsilon$ in Adam is efficient, and experimental results validate the fast convergence and good generalization performance of EAdam.

\subsection{Language Modeling}

\begin{figure}[htb]
  \centering
  \subfigure[1-layer LSTM]{
  \includegraphics[width=0.31\linewidth]{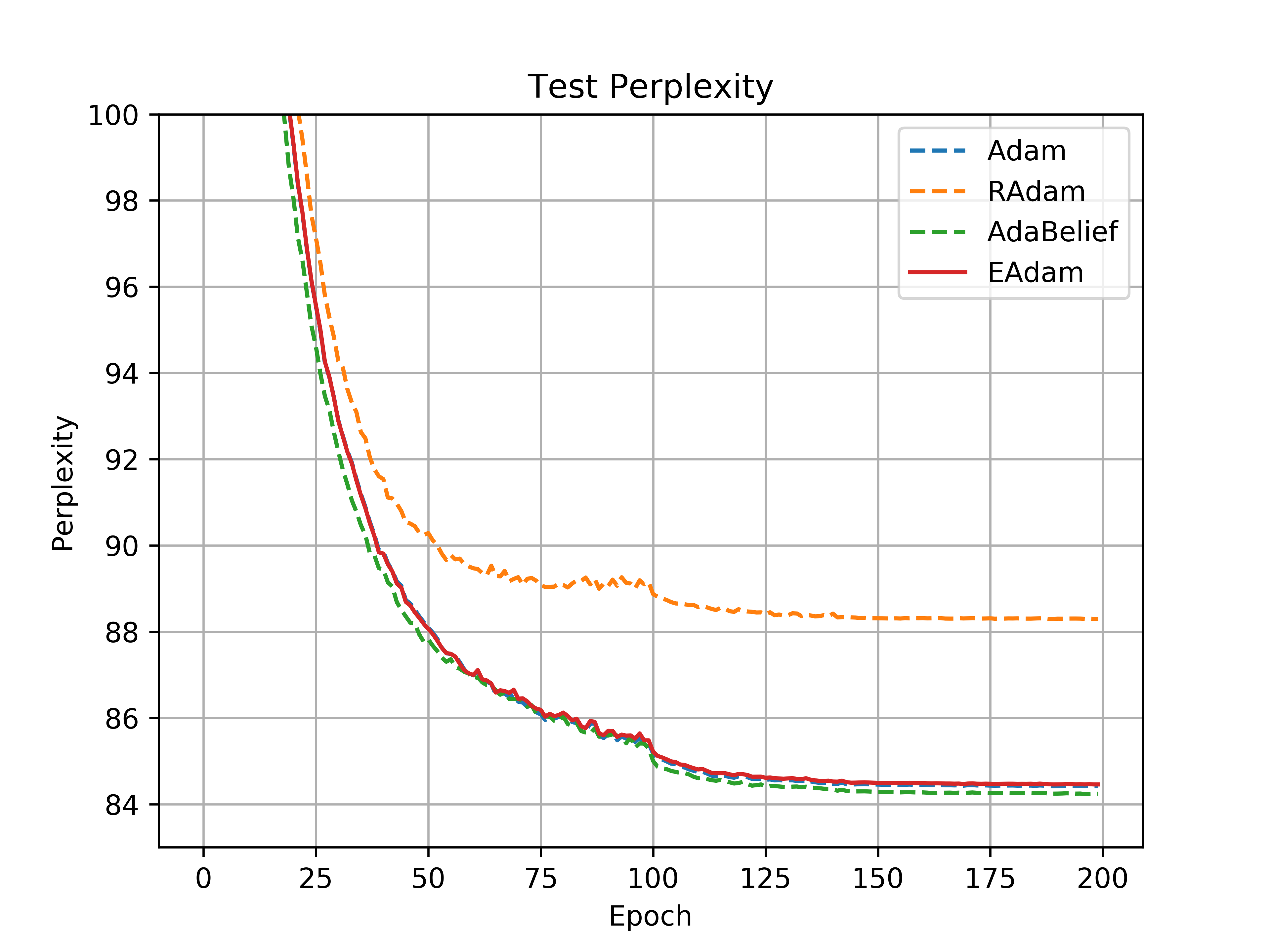}}
  \subfigure[2-layer LSTM]{
  \includegraphics[width=0.31\linewidth]{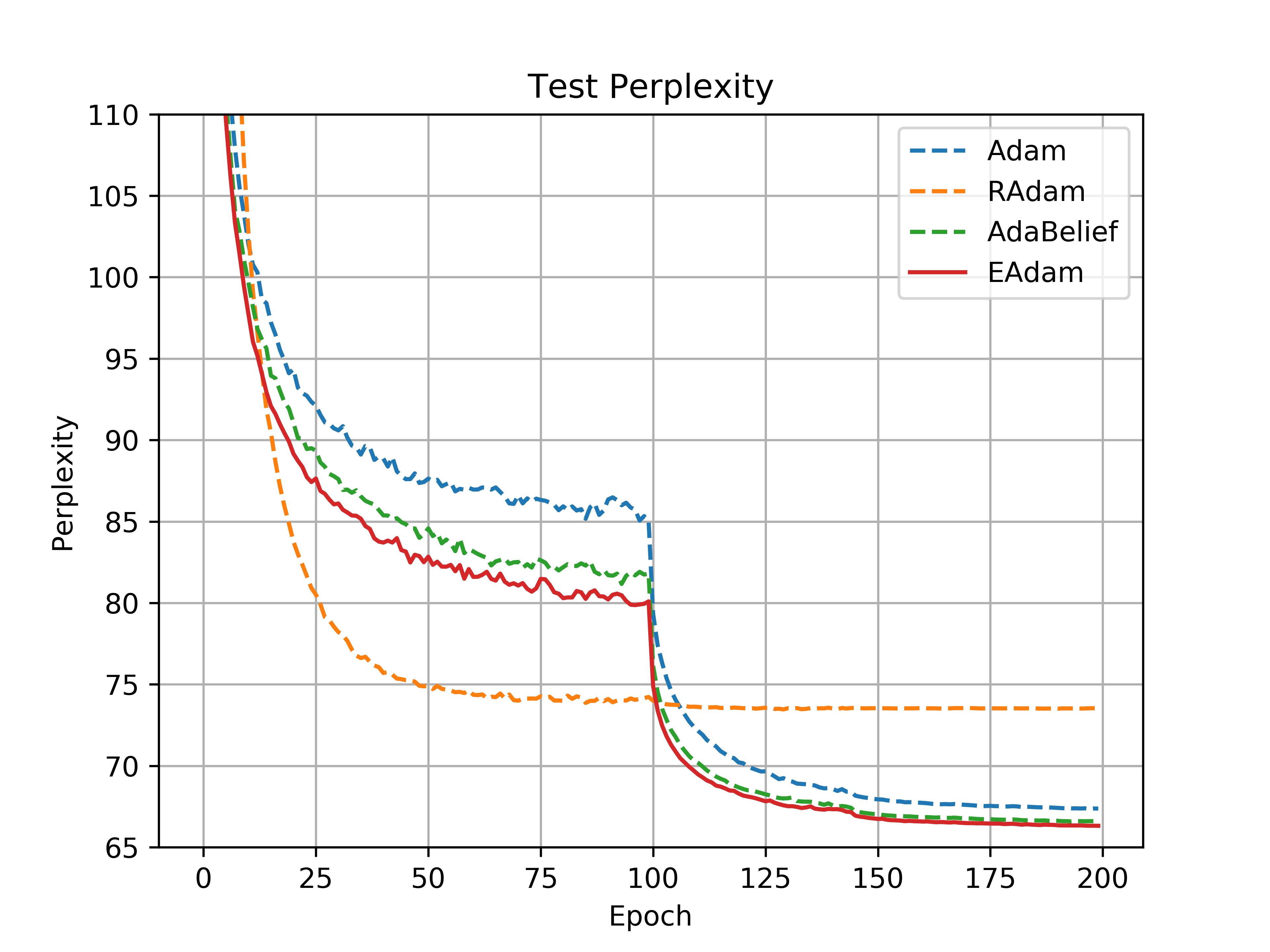}}
  \subfigure[3-layer LSTM]{
  \includegraphics[width=0.31\linewidth]{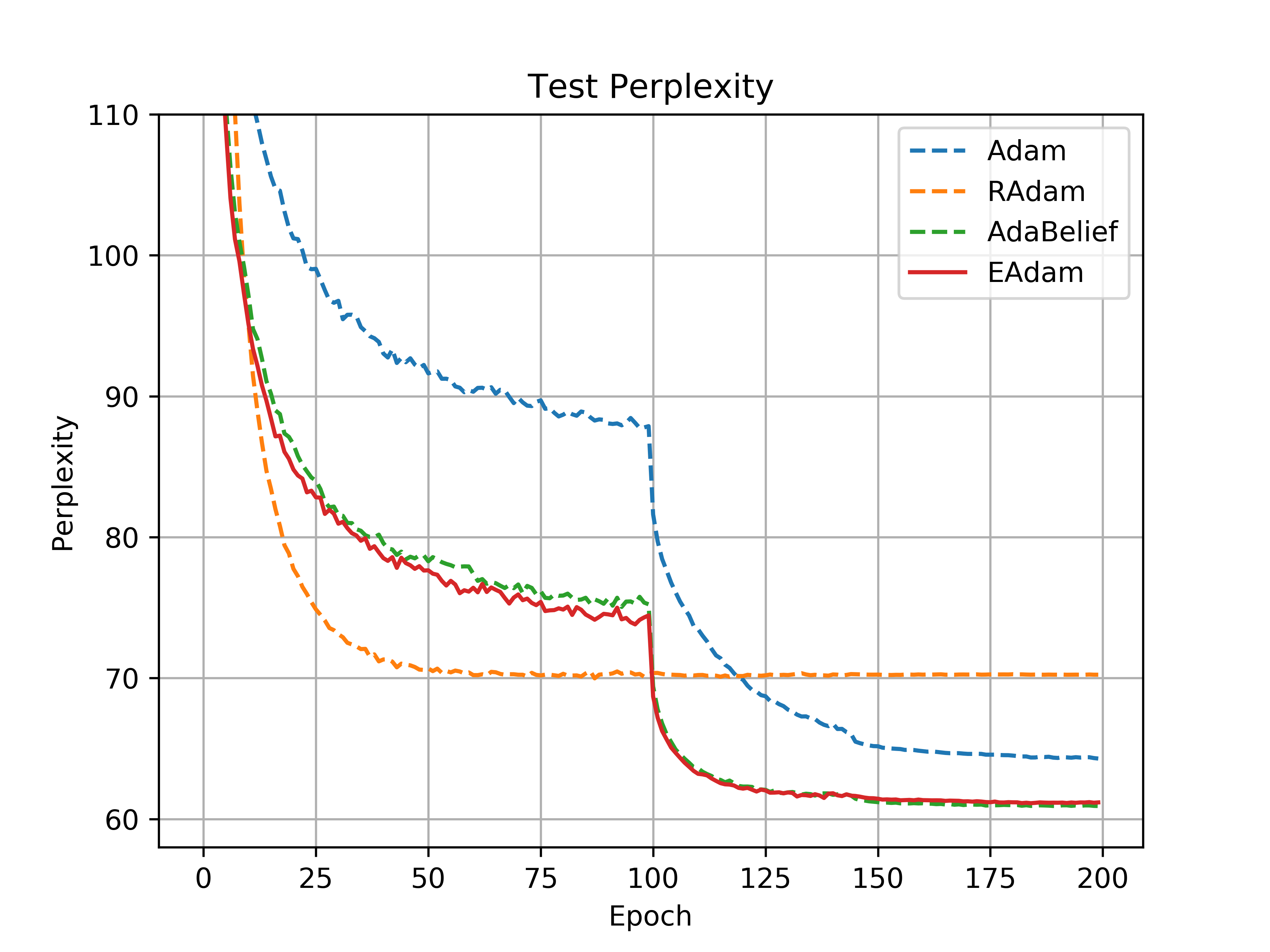}}\\
  \caption{The curves of test perplexity with epochs for Adam, RAdam, Adabelief and EAdam on Penn Treebank. The models we used here are 1,2,3-layer LSTM.}\label{picacc2}
\end{figure}

We next conduct an experiment on the language modeling task. We consider the LSTM network on the Penn Treebank dataset. For all experiments in this part, we set the weight decay as $5e-4$ of all methods and train all models for 200 epochs with batch-size 20 and multiply the learning rates by 0.1 at the 100-th epoch and 145-th epoch. Other related parameters are set as $\alpha\in\{10^{-4},10^{-3},10^{-2},10^{-1}\},\beta_1=0.9, \beta_2=0.999, \epsilon\in\{10^{-16},10^{-12},10^{-8}\}$. Results are summarized in Figure \ref{picacc2} and Table \ref{tabacc2}.

\begin{table}[htb]
\small
\setlength{\abovecaptionskip}{0.0cm}
\setlength{\belowcaptionskip}{0.3cm}
\centering
\caption{Test perplexity (lower is better) on Language Modeling for the 1,2,3-Layer LSTM on Penn Treebank .}
\label{tabacc2}
\begin{tabular}{@{}cc|ccccc@{}}
\toprule
Dataset   & Model    & Adam            & RAdam            & Adabelief           & EAdam           \\ \midrule
Treebank      & 1-Layer LSTM    & 84.42 & 88.30 & 84.24 & 84.46  \\
Treebank      & 2-Layer LSTM & 67.36 & 73.46 & 66.58 & 66.32  \\
Treebank  & 3-Layer LSTM   & 64.29 & 69.98 & 60.93 & 61.13  \\
 \bottomrule
\end{tabular}
\end{table}

Perplexity curves are shown in Figure \ref{picacc2}. We can see that RAdam performs worse than Adam, Adabelief and EAdam. For the 1-layer model, the performance of EAdam is similar to Adam and Adabelief. For the 2-layer and 3-layer LSTM models, EAdam outperforms Adam and has similar performance to Adabelief. As shown in Table \ref{tabacc2}, Adabelief achieves the lowest perplexity on the 1-layer and 3-layer LSTM model but the perplexity of EAdam is close to Adabelief, and EAdam achieves the lowest perplexity on 2-layer LSTM model. In a word, EAdam also can keep fast convergence and good accuracy in the language modeling task.

\subsection{Object detection}

\begin{figure}[htbp]
\centering
\subfigure[Adam]
{
    \begin{minipage}[b]{.24\linewidth}
        \centering
        \includegraphics[width=1\linewidth]{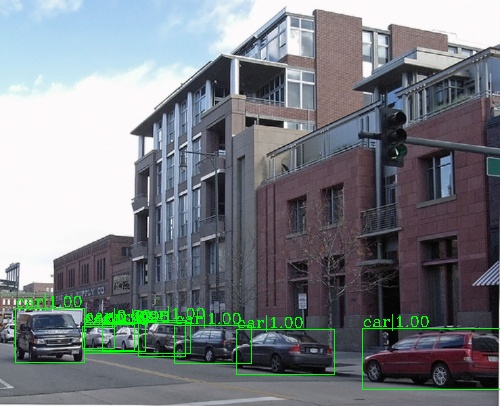} \\
        \includegraphics[width=1\linewidth]{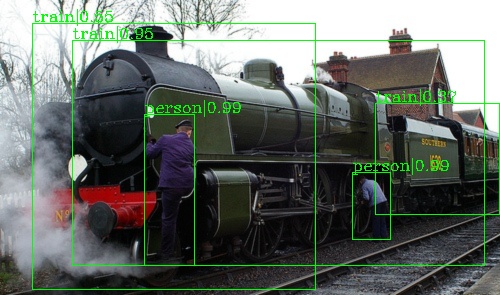}
    \end{minipage}
}\hspace{-3mm}
\subfigure[RAdam]
{
    \begin{minipage}[b]{.24\linewidth}
        \centering
        \includegraphics[width=1\linewidth]{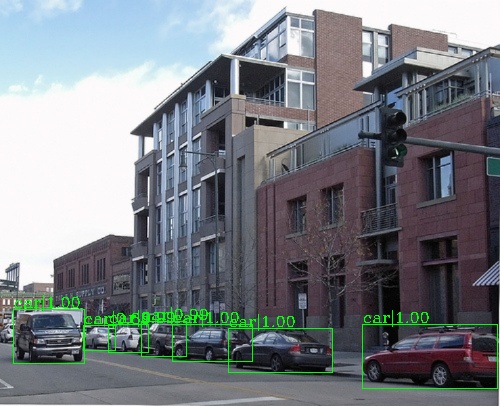} \\
        \includegraphics[width=1\linewidth]{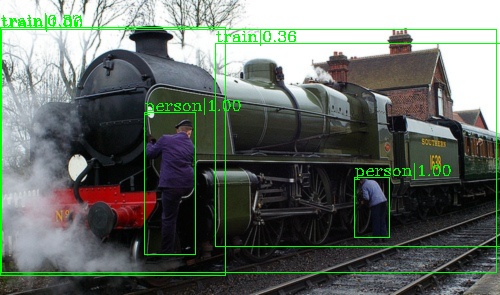}
    \end{minipage}
}\hspace{-3mm}
\subfigure[Adabelief]
{
    \begin{minipage}[b]{.24\linewidth}
        \centering
        \includegraphics[width=1\linewidth]{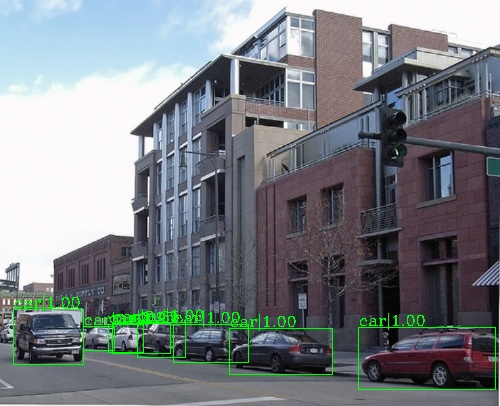} \\
        \includegraphics[width=1\linewidth]{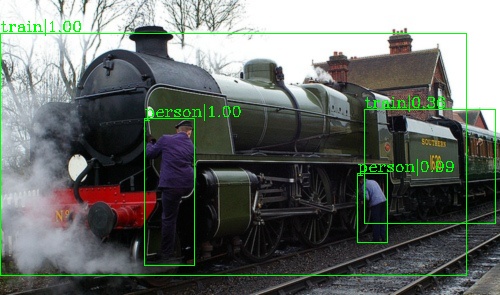}
    \end{minipage}
}\hspace{-3mm}
\subfigure[EAdam]
{
    \begin{minipage}[b]{.24\linewidth}
        \centering
        \includegraphics[width=1\linewidth]{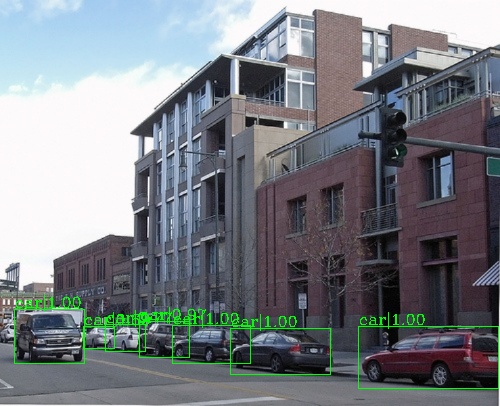} \\
        \includegraphics[width=1\linewidth]{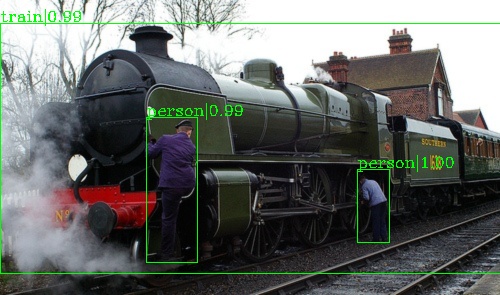}
    \end{minipage}
}
\caption{Detection examples using Faster-RCNN + FPN trained on PASCAL VOC.}\label{picacc3}
\end{figure}

Finally, we do object detection experiments on the PASCAL VOC dataset. The model we used here is pre-trained on ImageNet and the pre-trained model is from the official website. We train this model on the VOC2007 and VOC2012 trainval dataset (17K) and evaluate on the VOC2007 test dataset (5K). We use the MMDetection \citep{mm2019} toolbox as the detection framework. The official implementations and settings are used for all experiments. The model we used is Faster-RCNN + FPN. The backbone is ResNet50. We set the weight decay as $10^{-4}$ of all methods. We train the model for 4 epochs with batch-size 2 and multiply the learning rates by 0.1 at the last epoch. Other parameters are set as $\alpha=10^{-4},\beta_1=0.9, \beta_2=0.999, \epsilon=10^{-8}$ for all methods. Results are summarized in Figure \ref{picacc3} and Table \ref{tabacc3}. As we except, EAdam outperforms Adam, RAdam and has similar performance to AdaBelief. These results also illustrates that our method is still efficient in object detection tasks.

\begin{table}[htb]
\small
\setlength{\abovecaptionskip}{0.0cm}
\setlength{\belowcaptionskip}{0.3cm}
\centering
\caption{The mAP (higher is better) on PASCAL VOC using Faster-RCNN+FPN.}
\label{tabacc3}
\begin{tabular}{@{}c|ccccc@{}}
\toprule
   Method    & Adam            & RAdam            & Adabelief           & EAdam           \\ \midrule
   mAP    & 71.47 & 76.58 & 81.02 & 80.62  \\
 \bottomrule
\end{tabular}
\end{table}

\section{Conclusions}\label{sec-6}

In this paper, we discussed the impact of the constant $\epsilon$ in Adam and found that a simple change could bring significant improvement. So we proposed the EAdam algorithm, which is a new variant of Adam. To our knowledge, EAdam is the first variant of Adam based on the discussion of $\epsilon$. We also gave some intuitive explanations of the relationships and differences between EAdam and Adam. Finally, we validated the benefits of EAdam on extensive experiments. Of course, the exact reason why $\epsilon$ impact the performance of Adam greatly is still mysterious for us. What's more, can EAdam still outperform Adam on other DNNs or more complex large-scale training tasks, such as image classification on ImageNet and object detection on COCO? These deserve to be discussed and verified. This paper is only the first draft,  we will further enrich and improve the analysis and experiments in the next version.

\bibliographystyle{plainnat}
\bibliography{d}


\end{document}